\documentclass[letterpaper, 10 pt, conference]{ieeeconf}  

\IEEEoverridecommandlockouts                              

\usepackage{graphics} 
\usepackage{epsfig} 
\usepackage{mathptmx} 
\usepackage{times} 
\usepackage{amsmath} 
\usepackage{amssymb}  
\usepackage{todonotes}
\usepackage{subcaption}
\usepackage{multirow}
\usepackage{gensymb}
\usepackage{ifthen}
\usepackage{bm}
\usepackage{cite}
\usepackage{siunitx}
\usepackage{tabularx}
\usepackage{physics}
\AtBeginDocument{\RenewCommandCopy\qty\SI}
\makeatletter
\let\NAT@parse\undefined
\makeatother
\usepackage{hyperref}
\DeclareMathAlphabet{\mathcal}{OMS}{cmsy}{m}{n}
\newcommand{\dataset}[1]{\texttt{#1}}

\usepackage{fancyhdr}
\fancypagestyle{specialfooter}{%
  \fancyhf{}
  
  \fancyfoot[L]{\footnotesize This work was accepted by the 2025 IEEE International Conference on Robotics \& Automation, Atlanta, USA.\linebreak
  © 2025 IEEE. Personal use of this material is permitted.  Permission from IEEE must be obtained for all other uses, in any current or future media, including reprinting/republishing this material for advertising or promotional purposes, creating new collective works, for resale or redistribution to servers or lists, or reuse of any copyrighted component of this work in other works.}
}

\title{\LARGE \bf
Introspective Loop Closure for SLAM with 4D Imaging Radar
}

\author{
	\begin{tabular}{ccc}
		Maximilian Hilger$^{1}$ & Vladimír Kubelka$^{2}$ & Daniel Adolfsson$^{2}$  \\
		Ralf Becker$^{3}$  & Henrik Andreasson$^{2}$    & Achim J. Lilienthal$^{1,2}$ 
	\end{tabular}%
\thanks{This work was supported by the EU H2020 program (ITN project MORE, No. 858101), the MIRMI Seed Fund project DeepSTEP, and by the EU Horizon Europe Framework Programme (RaCOON project, ID: 101106906).}
\thanks{$^{1}$Technical University of Munich, Munich Institute for Robotics and Machine Intelligence (MIRMI), Chair of Perception for Intelligent Systems, Munich, Germany. Corresponding author:
        {\tt\small maximilian.hilger@tum.de}}%
\thanks{$^{2}$ Örebro University, AASS research centre, Robot Navigation and Perception Lab, \"Orebro, Sweden}%
\thanks{$^{3}$ Bosch Rexroth AG, Germany}%
}

\begin{document}

\maketitle
\thispagestyle{empty}
\pagestyle{empty}
\thispagestyle{specialfooter} 

\begin{abstract}
Simultaneous Localization and Mapping (SLAM) allows mobile robots to navigate without external positioning systems or pre-existing maps. 
Radar is emerging as a valuable sensing tool, especially in vision-obstructed environments, as it is less affected by particles than lidars or cameras. 
Modern 4D imaging radars provide three-dimensional geometric information and relative velocity measurements, but they bring challenges, such as a small field of view and sparse, noisy point clouds.
Detecting loop closures in SLAM is critical for reducing trajectory drift and maintaining map accuracy. 
However, the directional nature of 4D radar data makes identifying loop closures, especially from reverse viewpoints, difficult due to limited scan overlap.
This article explores using 4D radar for loop closure in SLAM, focusing on similar and opposing viewpoints. 
We generate submaps for a denser environment representation and use introspective measures to reject false detections in feature-degenerate environments. 
Our experiments show accurate loop closure detection in geometrically diverse settings for both similar and opposing viewpoints, improving trajectory estimation with up to \qty{82}{\percent} improvement in \emph{ATE} and rejecting false positives in self-similar environments.
\end{abstract}

\section{INTRODUCTION}
Robust SLAM is necessary to enable the reliable operation of mobile robots in harsh settings.
In this environment, dirt deposition on the sensor leads to visual obstruction, making established sensing modalities such as lidar and cameras failure-prone.
Radar sensors, in contrast, can penetrate the obstructions due to their longer signal wavelength.
Modern imaging radar sensors generate point clouds with up to 10000 points per scan, offering range, azimuth, and elevation measurements.
Rich four-dimensional data can be obtained with relative velocity measurements (further denoted as \emph{Doppler} measurements).
However, compared to 2D spinning radars, the point clouds of 4D radars are more noisy and suffer from a small field of view (FOV).
This makes tasks that require geometric information, such as scan matching and loop closure detection, challenging.
With the small FOV (typically up to 120$\degree$ in azimuth), point clouds showing the same scene from different viewpoints only share a small overlap.
In loop closure detection, this small overlap impairs both the retrieval and the metric alignment of loop closure candidates.
Especially when revisiting a location in reverse direction, detecting loops with opposing viewpoints is hard.

\begin{figure}
    \centering
	\begin{subfigure}[b]{0.125\textwidth}
		\includegraphics[trim={0, 0cm, 0, 0cm},clip,width = 1\textwidth]{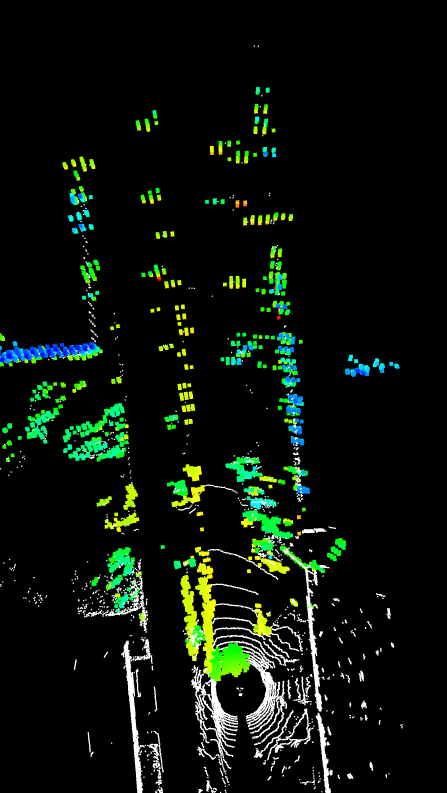}
		\caption{}
		\label{subfig:scan}
	\end{subfigure}
	\begin{subfigure}[b]{0.34\textwidth}
        \includegraphics[trim={0, 0cm, 0, 0cm},clip,width = 1\textwidth]{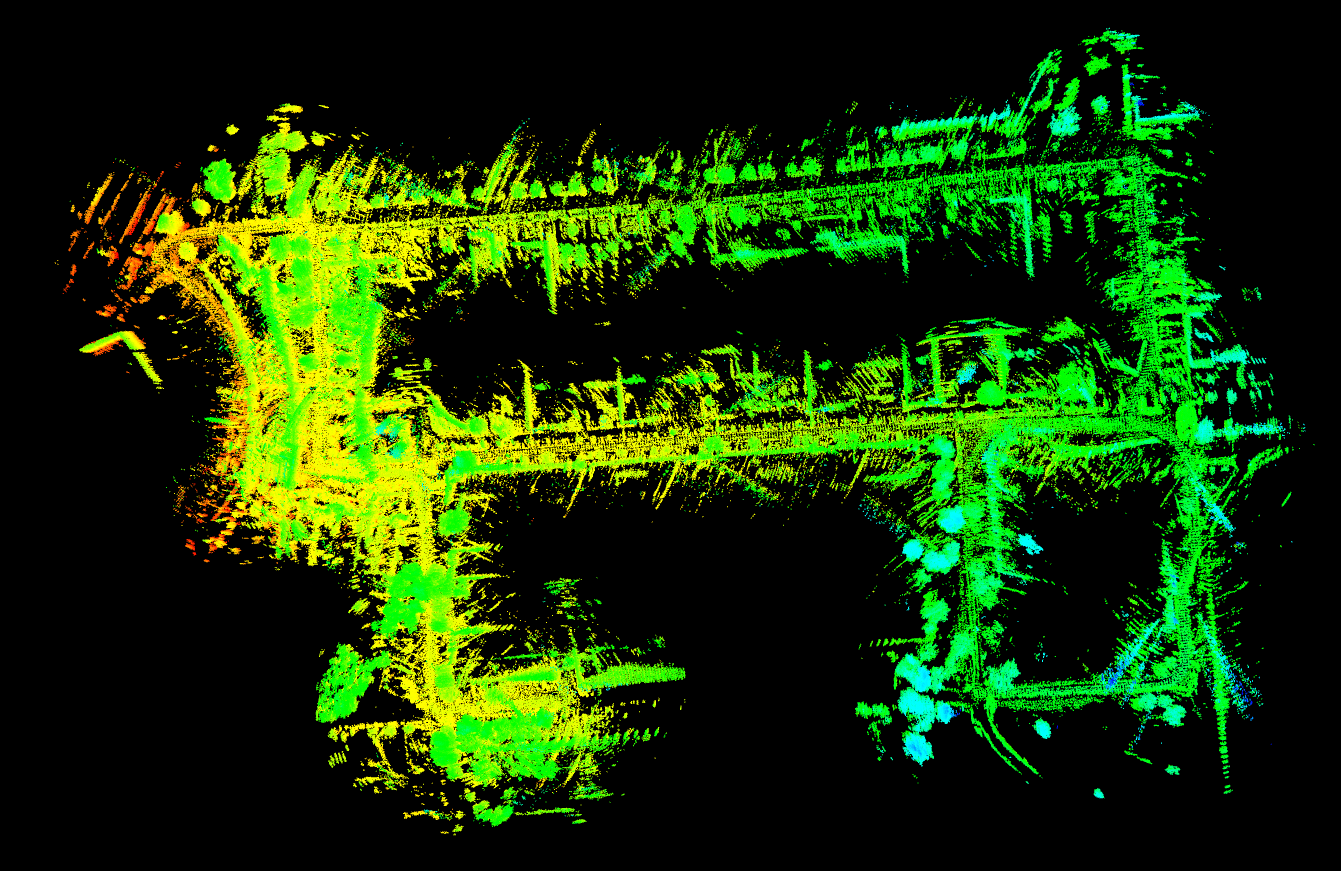}
		\caption{}
		\label{subfig:map}
	\end{subfigure}
    \caption{
    Our approach generates consistent maps from sparse 4D imaging radar data.
    (a) 4D radar (colored) and lidar (white) scans,
    (b) estimated map of the \dataset{Campus} environment}
    \label{fig:enter-label}
\end{figure}

Initial work on 3D SLAM with 4D radar \cite{Zhang.2023, Li.2023, Zhuang.2023, Wang.2023, Hilger.2024} applies the polar \emph{ScanContext} descriptor \cite{Kim.2018} for loop closure detection, which was originally developed for sensors with \qty{360}{\degree} coverage of the environment.
Using this approach, only a part of the descriptor is populated due to the limited FOV, and only loop closures with minor viewpoint variations can be detected \cite{Hilger.2024}.
Hence, a different method is necessary to detect loop closures under opposing viewpoints.

This article presents an approach for loop closure detection from similar and opposing viewpoints in 3D SLAM with a single 4D radar. 
For this, we explore combining different loop verification metrics, as successfully done with 2D radar \cite{Adolfsson.2023tbv}.
By this, we introspectively check the detected loop closures before including them in the SLAM estimate.
We evaluate the loop closure performance on the \dataset{Mine-and-Forest-radar-dataset} \cite{Kubelka.2024} recorded in underground and forest environments, and on sequences recorded on a university campus.
Our contributions are: 
\begin{itemize}
    \item A loop retrieval pipeline, combining coupled odometry/appearance similarity search with sequence-based filtering to retrieve loop closure candidates with similar and opposing viewpoints.
    \item A verification classifier adapted to 4D radar that considers odometry similarity, descriptor distances, and introspective alignment measures to reject incorrect loop closures. 
\end{itemize}
To the author's knowledge, this is the first work specifically addressing opposite-direction loop closure using 4D radar.
We demonstrate successful loop closure detection in geometrically diverse settings for similar and opposing viewpoints.

\section{RELATED WORK} 
\label{sec:sota}
Many modern SLAM systems rely on pose graph optimization. 
In these systems, an odometry estimator generates odometry edges, and place recognition followed by scan registration yields loop closure edges.
Optimizing the pose graph ensures consistent and accurate map and trajectory estimation.
Following, we present the related work in each of these components for 4D radar SLAM.

\subsection{Odometry with 4D imaging radar}
Existing radar odometry methods for 4D imaging radar can be grouped into instantaneous and correspondence-based methods.
Instantaneous approaches \cite{Doer.2020,Kramer.2020,Doer.2021,Ng.2021,Galeote.2023,Huang.2024,Kubelka.2024} leverage the Doppler velocity measured by the sensor to estimate the ego-motion of the sensor. The position is obtained by integrating the velocity over time. Here, the robot's orientation is recoverd using an IMU or kinematic assumptions. 
Correspondence-based odometry \cite{Retan.2021,Retan.2022,Michalczyk.2022,Michalczyk.2023,Lu.2024,Casado.2024} uses the spatial information of the radar points to establish correspondences to previously recorded maps or scans. 
Recently, Kubelka et al. \cite{Kubelka.2024} compared different radar odometry approaches. 
They conclude that highly accurate Doppler measurements supported by a low-drift IMU favor instantaneous approaches.
In contrast, correspondence-based methods work better for radars with longer target persistence in subsequent scans and when using low-cost IMUs.
In our work, we adopt the instantaneous Doppler+IMU approach from \cite{Kubelka.2024} for odometry estimation and submapping.

\subsection{Radar place recognition}
Most approaches to this day accomplish radar-based place recognition using spinning 2D radars, as these offer access to the full $360 \degree$ spectrum of the radar data. 
Here, either hand-crafted \cite{Hong.2020,Kim.2020,Jang.2023,Choi.2024,Gadd.2024} or learning-based approaches \cite{Saftescu.2020,Barnes.2020,Gadd.2020,Wang.2021,Komorowski.2021,Yuan.2023} are used. 
With the emergence of cheaper and smaller solid state radar sensors, the interest in place recognition with these sensors is increasing. 
Autoplace \cite{Cai.2022} removes dynamic points, then uses a CNN-based spatio-temporal encoder, and refines the retrieved candidates using radar cross-section (RCS) information. 
The approach is designed for automotive radars without elevation information. 
TransLoc4D \cite{Peng.2024} combines a CNN-based local feature extractor with a transformer network for global descriptor generation. 
Casado Herraez et al. \cite{Herraez.2024} use point convolutions for local feature encoding, point importance estimation, and RCS data to generate a small-sized descriptor. 
Their approach operates on single scans and does not rely on odometry information. 
Meng et al. \cite{Meng.2024} mount a single-chip radar on a rotating device to increase the covered FOV.
They compute descriptors using a spatial encoder on range-azimuth-heatmaps.
In our work, we employ the handcrafted \emph{CartContext} descriptor as part of the loop closure detection pipeline.

\subsection{Narrow FOV place recognition}
The challenge of the narrow field of view is not exclusive to 4D radar. 
Other sensing modalities, such as solid-state lidar and stereo cameras, have similar limitations. 
Recently, Kim et al. \cite{Kim.2024} proposed the SOLiD descriptor specifically tailored for solid-state lidar. 
They leverage the large vertical FOV of their sensors to form the descriptor. 
However, common 4D radars have a significantly smaller vertical opening angle, which makes this descriptor infeasible for 4D radar SLAM. 
Gupta et al.~\cite{Gupta.2024} accumulate point clouds in a voxel grid to create density maps. 
In the resulting density images, they match ORB features via RANSAC to obtain similar places. 
For stereo cameras, Carmichael et al. propose SPOT \cite{Carmichael.2024}.
They accumulate point clouds in a fixed radius around the camera to
generate CartContext descriptors \cite{Kim.2022}. 
By augmenting the query descriptor and using sequence matching \cite{Milford.2012} in viewpoint-dependent distance matrixes, they recognize places from both similar and opposing viewpoints.
Our approach adopts the voxel grid-based submapping from \cite{Gupta.2024} and sequence-based matching approach from SPOT.

\subsection{Radar SLAM with 4D imaging radar}
Several SLAM approaches have been proposed as a combination of radar odometry and radar place recognition. 
Zhang et al. \cite{Zhang.2023} estimate odometry with a variant of generalized ICP that incorporates the uncertainty in the measured radar points for scan-to-submap matching. 
They employ prior odometry information and barometric measurements to reduce the search space for loop closures, limiting their approach to viewpoint variations of up to \qty{20}{\degree}.
The feasible candidates are matched using intensity ScanContext \cite{Wang.2020}.
Li, Zhang, and Chen \cite{Li.2023} use scan-to-submap NDT matching to create radar odometry estimates. 
They obtain loop closures using ScanContext \cite{Kim.2018}, and fuse the loop closure factors together with odometry factors and preintegrated ego-velocity factors in a pose graph.
Zhuang et al. \cite{Zhuang.2023} combine a 4D radar with an IMU in an iterative extended Kalman filter for odometry estimation. 
Loop closures are detected using ScanContext \cite{Kim.2018} and generalized ICP matching. 
In their follow-up work \cite{Wang.2023}, they use multiple scans to form the descriptor. 
The authors report higher accuracy but do not quantify the improvement.
Similar to the existing approaches, we use a hand-crafted descriptor. However, we complement the appearance matching with further information and thoroughly evaluate the loop closure performance.

\section{METHODOLOGY}

\begin{figure*}[t!]
	\centering
	\includegraphics[trim={0, 0.5cm, 0, 0.4cm},clip,width = 0.93\textwidth]{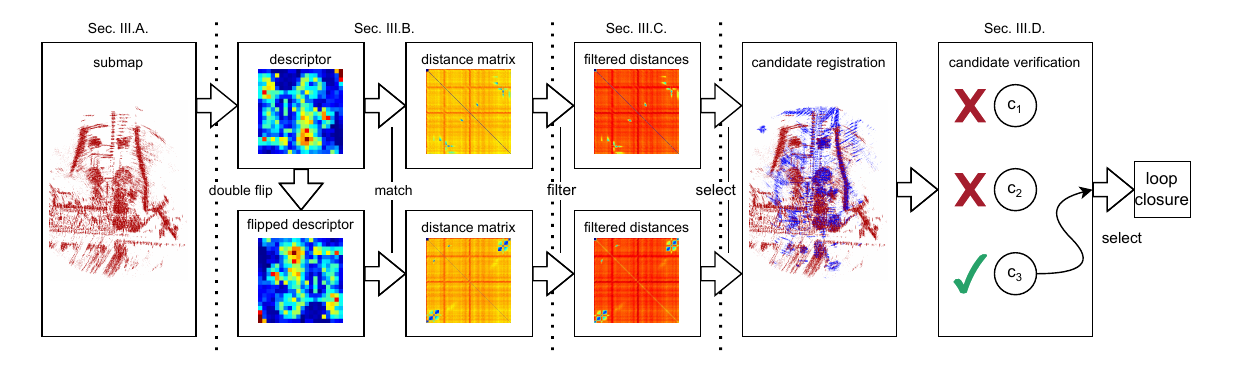}
	\caption{Loop retrieval and verification pipeline. Top row: loop closures with a similar viewpoint, bottom row: loop closures with an opposing viewpoint. Separating the viewpoints enables detecting loop closures in both directions}
	\label{fig:pipeline}
\end{figure*}
Our approach takes inspiration from TBV SLAM \cite{Adolfsson.2023tbv}, the current state-of-the-art in SLAM for spinning radar that focus on careful verification of loop closures. Our pipeline instead operates on point clouds from the more challenging 4D imaging radar and includes the following components:
\begin{enumerate}
    \item Odometry estimation and submapping (cf. Sec.~\ref{subsec:odom})
    \item Descriptor generation and matching (cf. Sec.~\ref{subsec:placerec})
    \item Sequence-based filtering (cf. Sec.~\ref{subsec:sequence})
    \item Computation and verification of loop candidates (cf. Sec.~\ref{subsec:verification})
    \item Pose graph optimization
\end{enumerate}
Fig.~\ref{fig:pipeline} gives an overview of our loop closure retrieval and verification approach.  

\subsection{Odometry estimation and submapping}
\label{subsec:odom}
We use the "Doppler+IMU" approach proposed in \cite{Kubelka.2024} to compute odometry. 
This method estimates the linear velocity of the sensor using three-point RANSAC on the measured Doppler velocities of the radar points \cite{Doer.2020}. 
The orientation obtained from an IMU is used to rotate the linear velocity into the world frame. 
Then, integrating the velocity yields the pose estimate. 
Interestingly, this method outperforms scan-matching-based approaches in the \dataset{Mine-and-forest-radar-dataset}.
During odometry estimation, we accumulate the static points extracted by RANCSAC in a robocentric voxel grid of resolution $\nu=\qty{1}{\meter}$.
We store a maximum of 20 points in each voxel to maintain a uniform point density and discard all points farther away than $r_a=\qty{50}{\meter}$ to ensure high run-time performance. 
We generate a new keyframe at a distance threshold $s = \qty{3}{\meter}$. 
For each keyframe, we store the current robocentric aggregated point cloud, which we refer to as \emph{submap}.

\subsection{Descriptor generation and matching} 
\label{subsec:placerec}
The submaps hold detailed information of each keyframe's environment. 
The goal of descriptor generation and matching is to compress the information aggregated in the submap to allow rapid loop closure search, accounting for both similar and opposing viewpoint cases.
To use all available sensor information, we leverage both appearance- and odometry-based features to express similarity between keyframes.

We employ the \emph{CartContext} descriptor \cite{Kim.2022}, which is based on a cartesian representation of the radar scan, to compress the submap's appearance. 
The descriptor $\mathbf{I}$ is formed using all points within a rectangle of size $r_{lo} \times r_{la}$. 
This rectangle is divided into $n_{lo} \times n_{la}$ cells, and an encoding function computes a scalar value based on the points in each cell.
The original work's encoding function sets the cell's value to the maximum height $z_{max}$ of all points. 
In radar sensing, the measured height is often subject to heavy noise \cite{Zhang.2023}. 
An improved encoding alternative is the maximum intensity $p_{max}$ that has demonstrated a high level of descriptiveness in previous work \cite{Wang.2020}. 
A challenge, however, is that the radar used in this work frequently returns ground points with intensity values similar to the reflections originating from non-ground targets. Accordingly, using the maximum intensity would discard the free-space information in front of the sensor. 
Instead, we adopt the encoding function proposed in \cite{Adolfsson.2023tbv}: The values of the descriptor are calculated as the sum of intensities of all points within a cell, divided by an occupancy-information-balancing weight (1000 in our experiments). 
Empty cells are set to $\mathbf{I}(i,j)=-1$. 
This encoding function enhances vertical structures such as walls and trees in the sensor's FOV, which form strong features. 

In the descriptor matching step, we extend the coupled odometry/appearance search from \cite{Adolfsson.2023tbv} to accommodate the limited FOV. 
The method narrows the search space by using odometry as a prior, in other words, we compute loop feasibility given the estimated odometry.
Our approach performs place matching separately for similar viewpoints and opposing viewpoints (further denoted by $(\cdot)_{sv}$ and $(\cdot)_{ov}$).  
We calculate the descriptor for matches with similar viewpoints $\mathbf{I}^q_{sv}$ directly from the query submap.
Double-flipping the original descriptor yields a descriptor $\mathbf{I}^q_{ov}$ suitable for opposing viewpoint matches \cite{Kim.2022}. 
We express similarity between query and candidate keyframes as joint distance metric 
\begin{align}
    \label{eq:coupled_dist}
    d^{q,c}_{joint} = 0.5 d_{cc}(\mathbf {I}^{q},\mathbf {I}^{c}) + d^{q,c}_{odom}.
\end{align}
Here, $d_{cc}(\mathbf {I}^{q},\mathbf {I}^{c})$ describes the cosine distance between query descriptor and candidate descriptor $\mathbf{I}^c$, and $d^{q,c}_{odom}$ penalizes large deviations from the odometry prior. 
The factor of $0.5$ ensures that both addends have a value between 0 and 1 and is adjustable according to expected odometry quality. 

We derive the odometry similarity measure $d^{q,c}_{odom}$ from \cite{Adolfsson.2023tbv}.
However, an additional orientation similarity measure complements the position similarity.
By that, we constrain our candidate search further using all available sensory information.
We denote the robot pose at a query keyframe as $\mathbf{x}^q \in SE(3)$ with translation vector $\mathbf{t}^q$ and rotation matrix $\mathbf{R}^q$, and similarly for the candidate.

To calculate the odometry-based distance, we model both translation and orientation drift as Gaussians
\begin{align}
    d^{q,c}_{odom} &= 1-p(\mathbf{x}^{q,c}_{loop} | \mathbf{v}^{q:c}_{odom}), \\
    p \left(\mathbf{x}^{q,c}_{loop} | \mathbf{v}^{q:c}_{odom}\right) &= \exp\left(-\frac{t_{err}^2}{2\sigma_{trans}^2}\right) \cdot \exp\left(-\frac{\theta_{err}^2}{2 \sigma_{rot}^2}\right),
\end{align}
where $\mathbf{x}^{q,c}_{loop}$ is a possible loop closure with and $\mathbf{v}^{q:c}_{odom}$ is the trajectory of the robot between candidate and query keyframe.
The translational error $t_{err}$ is normalized by the distance traveled between candidate and query $dist(\mathbf{v}^{c:q}_{odom})$:
\begin{align}
    t_{err} = \frac{\max(\norm{\mathbf{t}^q-\mathbf{t}^c}-\epsilon_{trans},0)}{dist(\mathbf{v}^{c:q}_{odom})}
\end{align}
We conservatively assume an odometry drift of \qty{4}{\percent}, setting $\sigma_{trans} = \num{0.04}$. 
In line with \cite{Adolfsson.2023tbv}, we set $\epsilon_{trans} = \qty{5}{\meter}$. 
The rotational error $\theta_{err}$ is calculated differently for similar and opposing viewpoints.
We expect a rotation between query pose and candidate pose of $\bm{\Delta}\mathbf{R}_{sv} = yaw(0\degree)$ for similar viewpoints and $\bm{\Delta}\mathbf{R}_{ov} = yaw(180\degree)$ for opposing viewpoints.
The scalar angular difference is calculated as
\begin{align}
    \Delta\theta = \arccos \left(0.5 \left( tr\left(\mathbf{R}^{q^{-1}} \mathbf{R}^c  \bm{\Delta}\mathbf{R}\right)-1\right)\right).
\end{align}
We allow for a deviation of up to $\epsilon_{rot} = 5 \degree$ from the odometry prior before starting penalizing the error, which yields
\begin{align}
    \theta_{err} = \max(\Delta\theta - \epsilon_{rot}, 0). 
\end{align}
The standard deviation $\sigma_{rot}$ tunes how restrictive the penalty is and is heuristically set to $3\degree$ in this work.
Two distinct distance matrices for similar and opposing viewpoints store the combined distance values (cf.~Fig.~\ref{fig:pipeline}).

\subsection{Sequence-based filtering}
\label{subsec:sequence}
With the descriptor matching step introduced in Sec.~\ref{subsec:placerec}, we can retrieve possible loop candidates. 
However, multiple loop closure candidates might have a similar appearance, leading to false positive candidates. 
We can, however, make use of sequential information to prevent retrieving false loop closures. 
Intuitively, neighbouring descriptors to both query and correct candidate should match well. Our method takes inspiration from~\cite{Carmichael.2024}.
As we equidistantly sample the trajectory, consistently matched sequences appear as diagonal lines with low distance values $d_{joint}$ in the distance matrices. 
Sequences with similar viewpoints appear with negative slopes, and sequences with opposing viewpoints have a positive slope in their respective distance matrices.
The filtered distance metric $d_{filtered}^{q,c}$ is calculated as the sum of the last $k$ combined descriptor distances along the positive and negative slopes in the respective distance matrices
\begin{align}
    d_{filtered,sv}^{q,c} = \frac{1}{w}\sum_{i=0}^{w-1} d_{joint}^{q-i,c-i} , ~~ d_{filtered,ov}^{q,c} = \frac{1}{w}\sum_{i=0}^{w-1} d_{joint}^{q-i,c+i},
\end{align}
and yields two distinct filtered distance matrices (cf.~Fig.~\ref{fig:pipeline}).
The sequence length parameter can be tuned according to the expected overlapping trajectory segments.
We search both filtered distance matrices for a total of $k$ candidates with the lowest filtered distance $d_{filtered}^{q,c}$.

\subsection{Computation and verification of loop candidates}
\label{subsec:verification}
As a result of the previous step, we obtain a set of roughly aligned loop closure candidates. We need to accurately compute the alignment between these, and introspectively verify the the final loop closure.
We register the retrieved loop candidates submaps to the query submaps. Specifically, we find the alignment that minimize a sum of point-to-distribution distances. ~\cite{Adolfsson.2023cfear}. 
The registration is initialized by an orientation estimate provided by the loop closure module in Sec.~\ref{subsec:placerec}. 
From the final iteration of registration, we obtain the registration cost $C_f$, the average size of the registered point sets $C_a$, and the number of correspondences $C_o$ as alignment quality measures to detect incorrectly registered loop constraints.
Finally, we carry out the introspective verification combining odometry similarity $d_{odom}$, descriptor distance $d_{cc}$ and the alignment quality metrics $C_f$, $C_a$ and $C_o$.
In contrast to earlier work \cite{Adolfsson.2023tbv,Hilger.2024}, we do not train the alignment classification separately based on odometry estimates but integrate the alignment metrics directly into the feature vector $\mathbf{X}^{q,c_k}_{loop}$.
The reason is that in the unsupervised training, all positive samples have a similar viewpoint, while we are interested in detecting misalignment with varying viewpoints.
Finally, we train a logistic regression classifier to decide whether a loop closure is accepted or discarded:
\begin{align}
    &y_{loop}^{q,c_{k}} = \frac{1}{1+e^{-{\bm {\Theta} }\mathbf {X}^{q,c_{k}}_{loop}}}, \:\:\text{s.t.}\:\: y_{loop}^{q,c_{k}}> y_{th}, \\ &\mathbf {X}^{q,c_{k}}_{loop} = [d_{odom} \: d_{cc} \: C_f \: C_a \: C_o \: 1]^{T}.
\end{align}
We use the retrieved loop candidates as training examples, assigning positive and negative labels using a registration error threshold of \qty{4}{\meter} and \qty{2.5}{\degree}.
The best scoring loop closure above the decision threshold $y_{th}$ is added to the pose graph.

\section{EVALUATION}

We proposed an approach for loop closure detection using 4D radar sensors. 
In this section, we evaluate three claims: (i) Loop closures can be detected and verified using multiple quality metrics, (ii) sequence-based filtering and odometry-coupled search improve the retrieval of loop candidates, and (iii) loop closures can be detected with both similar and opposing viewpoints. As the only open-source available SLAM approach \cite{Zhang.2023} is limited to viewpoint changes of up to \qty{20}{\degree} and the available place recognition methods \cite{Cai.2022,Meng.2024} do not support the radar's height dimension, a fair comparison to other approaches is hardly possible. Hence, we focus our evaluation on the effects of our components.

\subsection{Evaluation metrics}

The goal of loop closure detection is to identify a closure candidate that is metrically aligned to the scan. 
Hence, we need to account for failures in the registration.
In line with TBV~SLAM~\cite{Adolfsson.2023tbv}, we consider detected loops with translation errors smaller than \qty{4}{\meter} and rotational deviations of less than \qty{2.5}{\degree} as true positives.
If no loop is detected, we assume a potential loop closure could have been found between keyframes with a translational distance of up to $\qty{6}{\meter}$.

\begin{table}[t!]
	\centering
	\caption{Parameter settings}
	\begin{tabular}{|l | l | l |}
		\hline
		  \textbf{parameter} & \textbf{value} & \textbf{description} \\
        \hline
		  $s$  & \qty{3}{\meter} & keyframe spacing \\
        \hline
		  $\nu$ & \qty{1}{\meter} & voxel grid resolution  \\
		\hline
        $r_a$ & \qty{50}{\meter} & voxel grid radius \\
        \hline
        $r_{lo}$, $r_{la}$ & \qty{30}{\meter} & side length of descriptor \\
        \hline
        $n_{lo}$, $n_{la}$ & 20 & descriptor dimension \\
        \hline
        $\sigma_{trans}$, $\sigma_{rot}$ & 0.04, \qty{3}{\degree} & standard deviations odometry similarity \\
        \hline
        $\epsilon_{trans}$, $\epsilon_{rot}$ & \qty{5}{\meter}, 5{\degree} & deviation threshold \\
        \hline
        $w$ & 15 (\dataset{Mine}), & sequence length \\
        & 6 (other) & \\
        \hline
	\end{tabular}	
	\label{tab:parameters}    
\end{table}

\subsection{Experimental setup}
We evaluate our approach using the \dataset{Mine-and-forest-radar-dataset} \cite{Kubelka.2024}. 
It consists of two sequences, \dataset{Mine}, and \dataset{Forest}. 
While the \dataset{Forest} sequence was recorded traversing a large loop two times in the same direction, the \dataset{Mine} sequence also includes loop traversals in reverse direction. 
In addition to these sequences, we recorded two traversals in a snowy campus environment. 
In these sequences, most of the loop closures have opposing viewpoints.
Both sequences (\dataset{Campus1} and \dataset{Campus2}) share the same route but were recorded with different radar settings.
In particular, \dataset{Campus1} was recorded with the same short-range settings used in the \dataset{Mine-and-forest-radar-datatset} with a maximum range of \qty{42}{\meter}. In contrast, \dataset{Campus2} uses a longer maximum range of \qty{78}{\meter} at the cost of a lower range resolution.
The dataset was recorded using a Sensrad Hugin A3 imaging radar (FOV: $\qty{80}{\degree}\times\qty{30}{\degree}$) and an Xsens MTi-30 IMU. 
As ground truth we use HDL-graph-slam \cite{Koide.2019} with lidar data.
While for the \dataset{Forest} and \dataset{Campus} sequences, RTK-GNSS is available, there is substantial noise because of the treetops in the \dataset{Forest} and an underpass in the \dataset{Campus}.
We list our parameter settings in Tab.~\ref{tab:parameters}

We divide the evaluation in three distinct parts: (i) ablation study (Sec.~\ref{subsec:ablation}), (ii) evaluation of trajectory quality (Sec.~\ref{subsec:trajectory}), and (iii) generalization across environments and sensor settings (Sec.~\ref{subsec:generalization}).

\subsection{Ablation study}
\label{subsec:ablation}
We design an ablation study to quantify the contributions of our respective components:
\begin{enumerate}
    \item \emph{CartContext-scan}: We use one radar scan to generate the descriptor and its double-flipped counterpart
    \item \emph{CartContext-submap}: Instead of single radar scans, we adopt the voxel grid to generate the descriptors
    \item \emph{sequence}: Additionally, use the sequence-based filter to retrieve the best candidate
    \item \emph{odometry similarity} (translational): We use the translational component of the odometry as an additional metric for loop retrieval and verification
    \item \emph{odometry similarity} (translational + rotational): We add the rotational component in the odometry similarity
    \item \emph{top 3 candidates}: We retrieve the 3 best candidates and select the candidate with the highest classifier score
\end{enumerate}

We conduct our ablation study separately in the \dataset{Mine} and \dataset{Campus1} sequences, which offer ample possibilities for loop closures with both similar and opposing viewpoints.
The \dataset{Mine} sequence shows a more self-similar environment than the \dataset{Campus} sequence.
To evaluate the different combinations, we employ precision-recall curves with
Precision~$P= \frac{\#TP}{\#TP + \#FP}$ and recall~$R= \frac{\#TP}{\#GTP}$ \cite{Schubert.2023}. 
Here, $\#TP$, $\#FP$, and  $\#GTP$ denote the number of true positives, false positives, and  ground truth positives, respectively. 

\begin{figure}
    \centering    
	\begin{subfigure}[b]{0.23\textwidth}
		\includegraphics[trim={0.2cm, 0.3cm, 0.2cm, 0.2cm},clip,width = 1\textwidth]{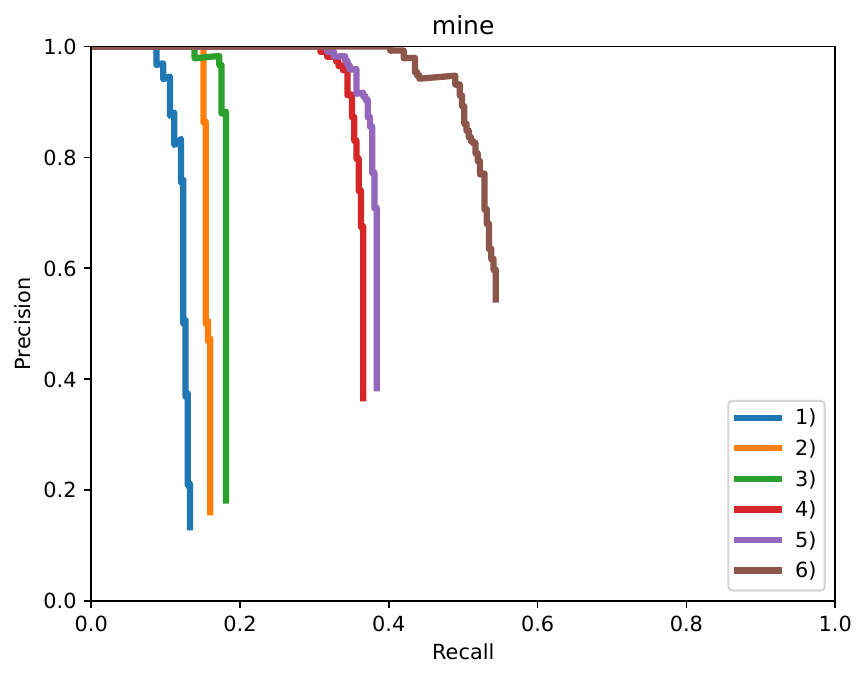}
		\caption{\dataset{Mine}}
		\label{subfig:ablation_mine}
	\end{subfigure}
	\begin{subfigure}[b]{0.23\textwidth}
		\includegraphics[trim={0.2cm, 0.3cm, 0.2cm, 0.2cm},clip,width = 1\textwidth]{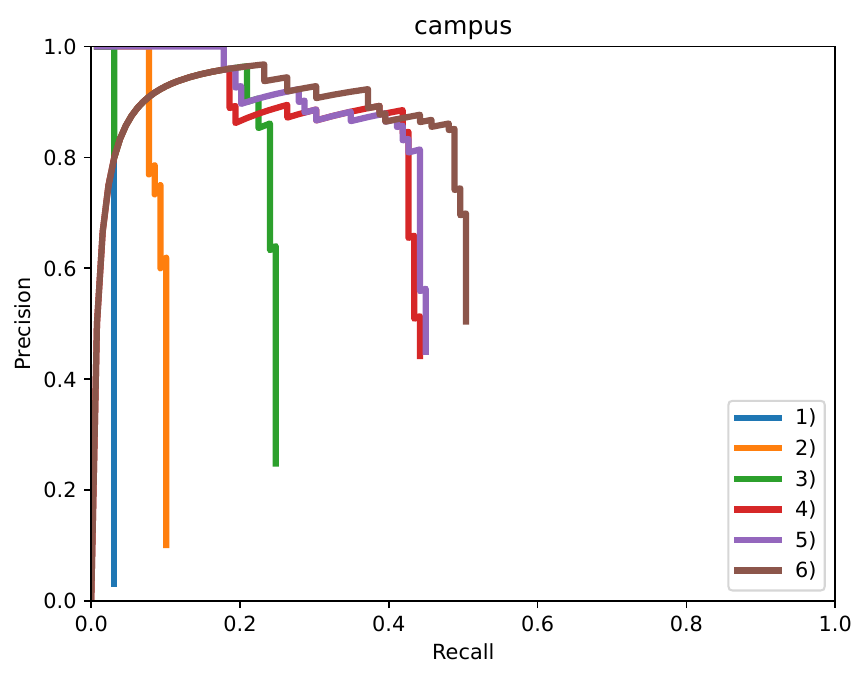}
		\caption{\dataset{Campus1}}
		\label{subfig:ablation_campussr}
	\end{subfigure}
    \caption{Precision-Recall curves of the ablation study}
    \label{fig:ablation}
\end{figure}

\begin{figure}
    \centering    
	\begin{subfigure}[b]{0.14\textwidth}
		\includegraphics[trim={0, 0cm, 0, 0cm},clip,width = 1\textwidth]{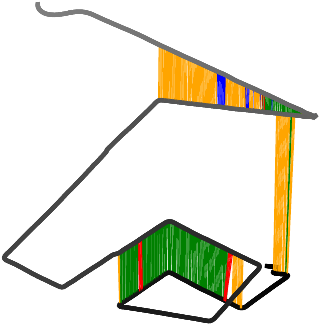}
		\caption{\dataset{Mine}}
		\label{subfig:closures_mine}
	\end{subfigure}
	\begin{subfigure}[b]{0.175\textwidth}
		\includegraphics[trim={0, 0cm, 0, 0cm},clip,width = 1\textwidth]{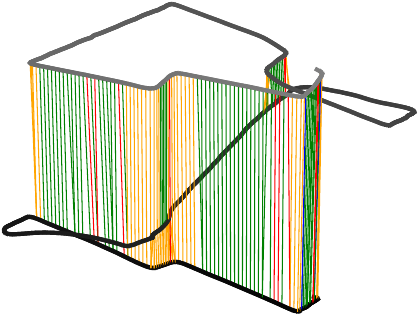}
		\caption{\dataset{Campus1}}
		\label{subfig:closures_campussr}
	\end{subfigure}
    \caption{Loop closures with full pipeline.
     Red (dangerous failure – false loop with undesired high confidence), orange (safe failure – false loop but with desired low confidence), blue (safe failure, correct loop with undesired low confidence), green (success, correct loop with desired high confidence). Decision threshold set according to the highest F1 value}
    \label{fig:closures}
\end{figure}

Fig.~\ref{fig:ablation} depicts the precision-recall curves in both sequences.
We see that both using submaps and sequence filtering moderately improve the performance in the \dataset{Mine} environment.
The effect is more prominent in the \dataset{Campus} environment. 
We theorize that additional radar points in the geometrically uniform mine do hardly yield additional structures in the descriptor.
Using the odometry-coupled search and verification substantially improves the achievable recall in both environments.
However, there is no pronounced benefit in using an orientation prior in the odometry similarity measure.
Hence, we conclude that selecting between same- and opposing-viewpoint candidates can be accomplished even without a prior orientation guess.
Finally, we observe that retrieving multiple loop closure candidates further improves the recall.
A possible risk occurs in the \dataset{Campus1} sequence: The highest-scoring candidate is false positive, as it can be observed in Fig.~\ref{subfig:ablation_campussr} where the precision-recall curve starts at zero precision.
Retrieving multiple candidates may also increase the risk of introducing false positive detections.

\begin{figure}
    \centering
	\begin{subfigure}[b]{0.3\textwidth}
 		\includegraphics[trim={0, 0.3cm, 0, 0.3cm},clip,width = 1\textwidth]{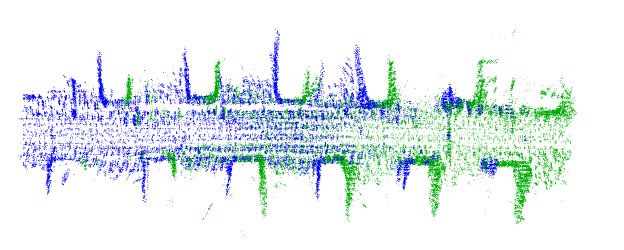}
		\caption{Query submap (green) correctly aligned to candidate submap (blue)}
		\label{subfig:loopcorrect}
	\end{subfigure}
	\begin{subfigure}[b]{0.3\textwidth}
 		\includegraphics[trim={0, 0.3cm, 0, 0.3cm},clip,width = 1\textwidth]{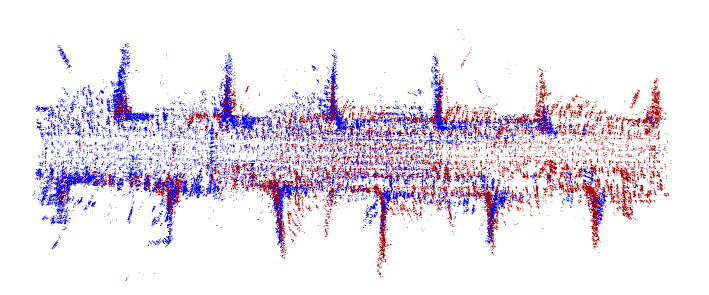}
		\caption{Query submap (red) incorrectly aligns to opposing walls in candidate submap}
		\label{subfig:wrongloops}
	\end{subfigure}
    \caption{Loop closure detection with side tunnels}
    \label{fig:oppviewinmine}
\end{figure}

We visualize the detected loop closures in both sequences with the decision threshold $y_{th}$ set such that it maximizes the F1-score in Fig.~\ref{fig:closures}.
Here, we can see that in both sequences, the loop closures in the same direction are almost always detected, and most of the opposite direction loop closures.
However, we frequently miss loop closures at corners.
This is a drawback of our design decision to focus on similar and opposing viewpoints. 
In corners, the orientation difference is too large for descriptors and point clouds to be successfully matched.
Additionally, we miss most loop closures in the long straight corridor in the \dataset{Mine} sequence. 
A possible reason for this is the encountered self-similar environment: On both sides of the tunnel, side tunnels are spaced evenly. 
The walls of the side tunnels make up strong features in the descriptor. 
If revisited from different directions, the two sides of the wall are incorrectly matched in the descriptor, as depicted in Fig.~\ref{fig:oppviewinmine}.
In this section, the submaps cannot be reliably matched, as no overlapping structures constrain the registration along the direction of travel.
However, both failure cases are mostly detected using our loop classifier, as shown by the orange color of the closures in Fig.~\ref{subfig:closures_mine}.
This example underlines the necessity for introspective approaches to ensure reliable and robust mapping.

\subsection{Evaluation of trajectory quality}
\label{subsec:trajectory}

\begin{figure}
    \centering
    
	\begin{subfigure}[b]{0.18\textwidth}
		\includegraphics[trim={0.1cm, 0.0cm, 0.1cm, 0cm},clip,width = 1\textwidth]{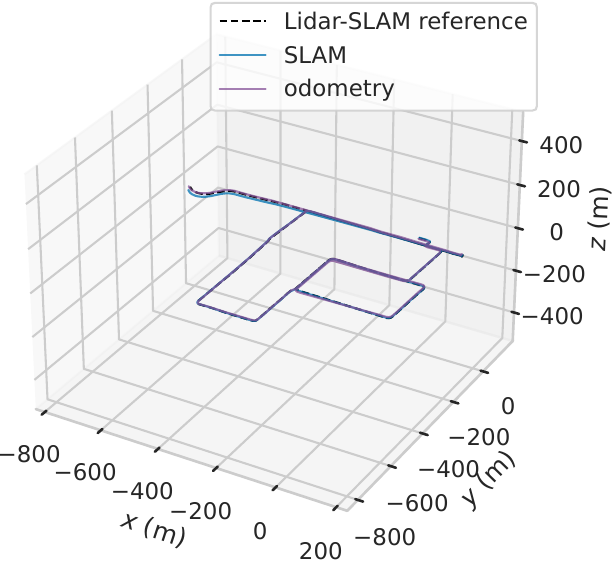}
		\caption{\dataset{Mine}}
		\label{subfig:trajmine}
	\end{subfigure}
	\begin{subfigure}[b]{0.18\textwidth}
		\includegraphics[trim={0.1cm, 0.0cm, 0.1cm, 0cm},clip,width = 1\textwidth]{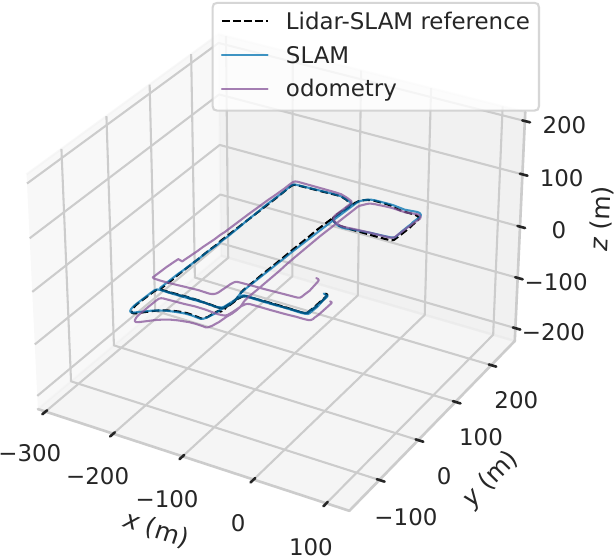}
		\caption{\dataset{Campus1}}
		\label{subfig:trajcampus}
	\end{subfigure}
    \caption{Trajectories on training sequences}
    \label{fig:traj}
\end{figure}

We use the KITTI odometry metric \cite{Geiger.2012} and the Absolute Trajectory Error (\textit{ATE}) to score the accuracy of the estimated trajectories.
The KITTI metric calculates the average relative error over segments of length \qtyrange{100}{800}{\meter} and yields the translational and rotational drift $t_{rel}$ and $r_{rel}$. 
The \textit{ATE} measures the Root Mean Squared Error between predicted poses and ground truth.
To underline the importance of loop closures, we include a comparison to the odometry without loop closure (cf. Tab.~\ref{tab:trajectorymetrics}). 
We observe improvements in translational drift and \emph{ATE} in both the \dataset{Mine} and the \dataset{Campus1} sequence.
The detected loop closures eliminate most of the drift in z-direction occurring in the \dataset{Campus1} sequence, leading to an improvement in \emph{ATE} of \qty{82}{\percent}.

\begin{table}[t!]
	\centering
	\caption{Trajectory performance metrics.}
	\begin{tabular}{| c | c c c | c c c |}
		\hline
		  ~  & \multicolumn{3}{c|}{radar odometry only} & \multicolumn{3}{c|}{full SLAM (ours)} \\
		  Metric  & $t_{rel}$ & $r_{rel}$  & \textit{ATE} & $t_{rel}$ & $r_{rel}$ & \textit{ATE} \\
		  Sequence  & [\%] & [\degree/100m] & [m] & [\%] & [\degree/100m] & [m] \\
		\hline
	    \dataset{Mine}       & 1.04 & 0.313 & 6.541 & 0.71 & 0.264 & 2.995 \\
        \dataset{Campus1} & 3.94 & 1.281 & 14.14 & 2.70 & 1.566 & 2.503 \\
        \dataset{Campus2} & 4.93 & 1.227 & 27.27 & 3.77 & 1.494 & 10.24 \\
        \dataset{Forest}     & 1.24 & 0.625 & 5.440 & 0.40 & 0.266 & 0.433 \\
		\hline
	\end{tabular}	
	\label{tab:trajectorymetrics}    
\end{table}

\subsection{Generalization across environments and sensor settings}
\label{subsec:generalization}

\begin{figure}
    \centering
	\begin{subfigure}[b]{0.18\textwidth}
		\includegraphics[trim={0.1cm, 0.0cm, 0.1cm, 0cm},clip,width = 1\textwidth]{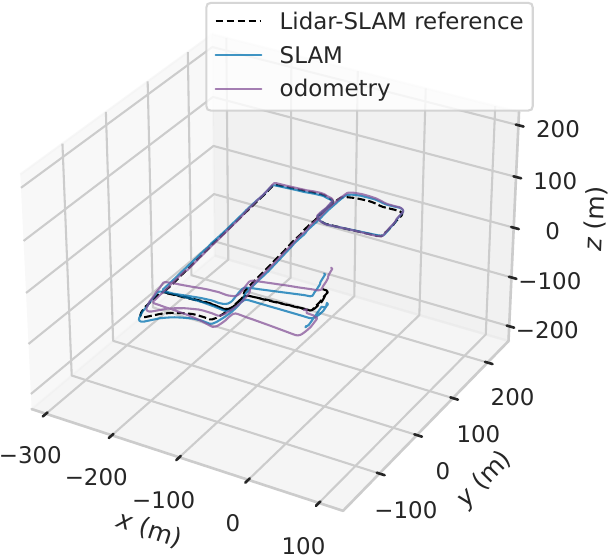}
		\caption{\dataset{Campus2}}
		\label{subfig:campus_trajectory}
	\end{subfigure}
	\begin{subfigure}[b]{0.17\textwidth}
		\includegraphics[trim={0.1cm, 0.0cm, 0.1cm, 0cm},clip,width = 1\textwidth]{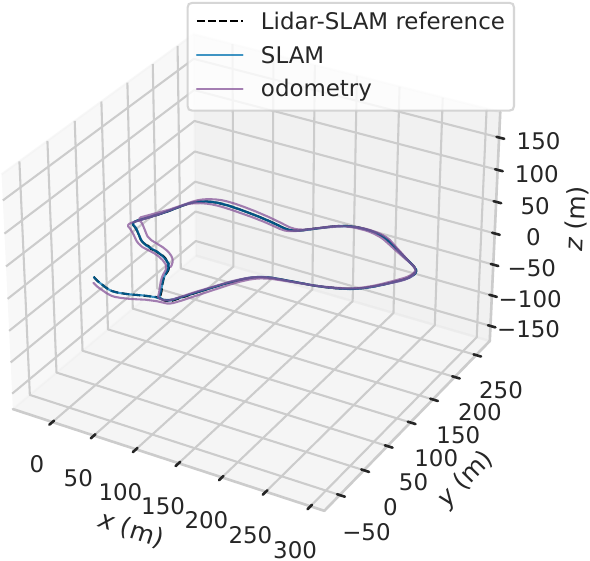}
		\caption{\dataset{Forest}}
		\label{subfig:forest_trajectory}
	\end{subfigure}
    \caption{Trajectories with classifier trained on \dataset{campus1}}
    \label{fig:generalization_trajectory}
\end{figure}

We use the classifier trained on the \dataset{Campus1} sequence and deploy it to the \dataset{Campus2} and \dataset{Forest} sequences.
In both sequences, drift and \emph{ATE} are substantially reduced with the detected loop closures (cf.~Tab.~\ref{tab:trajectorymetrics}). 
However, some loop closures in the \dataset{Campus2} are missed.
This shows that the trained classifier may be transferred to different environments but requires retraining for different point cloud characteristics.
Fig.~\ref{fig:generalization_trajectory} depicts the estimated trajectories.

\subsection{Run-time performance}
We evaluate the runtime performance in the \dataset{Mine} sequence.
Our measurements (with an AMD Ryzen 9 7950X CPU) are as follows:
Computation of surface normals: 28 ms/keyframe;
Descriptor generation: 0.8 ms/keyframe;
Loop retrieval: 8 ms/keyframe;
Registration: 13 ms/candidate; 
Verification 2 ms/candidate;
Loop closure: 86 ms @ k = 3;
Pose graph optimization (all constraints active with odometry as prior): 111 ms @ 1369 keyframes.
We conclude that our system in the considered environment runs at a sufficient speed for loop closure detection in a background thread of an online SLAM system.

\section{CONCLUSIONS}

In this paper, we presented a method for detecting and verifying loop closures with both similar and opposing viewpoints using a single 4D imaging radar.
We showed how the adoption of sequence-based filtering and matching can alleviate challenges inherited from the sensor's narrow FOV.
Additionally, we leveraged introspective verification to increase robustness and reject false loops, even in geometrically self-similar environments.
Experiments showed that the increased robustness leads to more accurate trajectory estimation and consistent mapping.
Our work paves the way toward robust and fail-safe SLAM for autonomous vehicles in harsh conditions.
Future work will aim to relax the viewpoint assumptions by using rotation invariant descriptors.

\addtolength{\textheight}{-0cm}   






\bibliographystyle{IEEEtran}
\bibliography{IEEEabrv,References.bib}

\end{document}